\documentclass{article}
\usepackage[nonatbib,preprint]{neurips_2022_modified}
\usepackage[numbers,sort&compress]{natbib}

\usepackage{subcaption}

\usepackage{url}
\usepackage{hyperref}

\usepackage{dsfont}
\usepackage{multirow}
\usepackage{booktabs}
\usepackage{colortbl}

\usepackage{wrapfig}
\usepackage{subcaption}
\usepackage{etoolbox}
\usepackage{verbatim}
\usepackage{amssymb}
\usepackage{mathrsfs}
\usepackage{bm}
\usepackage{algcompatible}
\usepackage[noend]{algpseudocode}
\usepackage{mathtools}
\usepackage{nccmath}
\usepackage{float}
\usepackage{times}
\usepackage{multirow,color,graphicx}
\usepackage{subcaption}
\usepackage{caption}
\usepackage{xfrac,amsbsy}
\usepackage[titletoc]{appendix}
\usepackage{xspace}
\usepackage{savesym,verbatim}
\usepackage{paralist}
\usepackage{subscript}
\usepackage{tikz}
\usepackage{bbm}
\usepackage{framed}

\usetikzlibrary{shapes.geometric}
\usetikzlibrary{shapes.misc}
\usetikzlibrary{decorations.pathreplacing}
\usetikzlibrary{patterns}
\usetikzlibrary{arrows,positioning,shapes}
\usetikzlibrary{arrows.meta}
\usepackage{picture}
\tikzset{cross/.style={cross out, draw=black, fill=none, minimum size=2*(#1-\pgflinewidth), inner sep=0pt, outer sep=0pt}, cross/.default={2pt}}
\usepackage{xcolor}
\usepackage{environ}
\usepackage{enumerate}
\usepackage{esvect} 
 
\DeclareMathOperator*{\argmin}{argmin} 
\usepackage[ruled,vlined]{algorithm2e}
\usepackage{optidef}
\usepackage{booktabs}
\usepackage{hhline}
\usepackage{makecell}
\usepackage{longtable}
\usepackage{mdframed}
\usepackage{siunitx}

\NewEnviron{boxcode2col}[4]{
    \begin{center}
   	 	\scalebox{#3}{
	   	 	\setlength\tabcolsep{4pt}
	   	 	\renewcommand{\arraystretch}{#4}
	   	 	\begin{tabular}{|p{#1}p{#2}|}   
    			\hline
				\BODY
   			 	[1ex]
    			\hline
    		\end{tabular}
    	}
    \end{center}
}

\NewEnviron{boxcode}[3]{
    \begin{center}
   	 	\scalebox{#2}{
	   	 	\setlength\tabcolsep{4pt}
	   	 	\renewcommand{\arraystretch}{#3}
    		\begin{tabular}{|p{#1}|}    
    			\hline
    			\vspace{0.1mm}    
				\BODY
    			\\\hline
    		\end{tabular}
    	}
    \end{center}
}

\newcommand{\taskspace}{\ensuremath{\mathbb{T}}}
\newcommand{\codespace}{\ensuremath{\mathbb{C}}}
\newcommand{\sketchspace}{\ensuremath{\mathbb{S}}}
\newcommand{\sketchmap}{\ensuremath{\Psi}}
\newcommand{\solutioncode}{\ensuremath{C^{\text{in,}\star}}}
\newcommand{\studentcode}{\ensuremath{C^{\text{in,stu}}}}
\newcommand{\exposureparam}{\ensuremath{k}}
\newcommand{\hintpolicy}{\ensuremath{\footnotesize{\texttt{GetSketch()}}}}

\newcommand{\algmincode}{\textsc{PQS-RedCode}}
\newcommand{\algsame}{\textsc{PQS-FullHop}}
\newcommand{\alghintpolicystruct}{\textsc{PQS-OneHop}}
\newcommand{\algmultihop}{\textsc{PQuizSyn}}
\newcommand{\substructures}{\textsc{SubStructs}}
\newcommand{\reducedcodes}{\textsc{RedCodes}}
\newcommand{\algnohint}{\textsc{NoHint}}
\newcommand{\algactionhint}{\textsc{NextStep}}
\newcommand{\textcode}[1]{{\fontfamily{cmtt}\selectfont #1}\xspace}

\newcommand{\DSLMove}{\textcode{move}}

\newcommand{\DSLTurnLeft}{\textcode{turnLeft}}

\newcommand{\DSLTurnRight}{\textcode{turnRight}}

\newcommand{\DSLPickMarker}{\textcode{pickMarker}}

\newcommand{\DSLPutMarker}{\textcode{putMarker}}
\newcommand{\DSLRepeat}{\textcode{\textsc{Repeat}}}
\newcommand{\DSLRepeatUntil}{\textcode{\textsc{RepeatUntil}}}
\newcommand{\DSLRUntil}{\textcode{\textsc{RUntil}}}
\newcommand{\DSLIf}{\textcode{\textsc{If}}}
\newcommand{\DSLIfElse}{\textcode{\textsc{IfElse}}}
\newcommand{\DSLElse}{\textcode{\textsc{Else}}}

\newcommand{\DSLWhile}{\textcode{\textsc{While}}}
\newcommand{\DSLRun}{\textcode{\textsc{Run}}}
\newcommand{\DSLBoolGoal}{\textcode{goal}}
\newcommand{\DSLBoolPathAhead}{\textcode{pathAhead}}

\newcommand{\DSLBoolPathRight}{\textcode{pathRight}}

\newcommand{\DSLBoolNoMarker}{\textcode{noMarker}}

\newcommand{\SDSLBool}{\textcode{B}}

\newcommand{\ind}{\emph{Adaptive}}
\newcommand{\interp}{\emph{Comprehensible}}
\newcommand{\conceal}{\emph{Concealing}}
\newcommand{\overall}{\emph{Overall}}

\title{Adaptive Scaffolding in Block-Based Programming\\ via Synthesizing New Tasks as Pop Quizzes}

\author{
  \textbf{Ahana Ghosh}\textsuperscript{1} \quad 
  \textbf{Sebastian Tschiatschek}\textsuperscript{2} \quad 
  \textbf{Sam Devlin}\textsuperscript{3} \quad
  \textbf{Adish Singla}\textsuperscript{1}\\
   \textsuperscript{1}MPI-SWS, \textcode{\{gahana, adishs\}@mpi-sws.org}\\
  \textsuperscript{2}University of Vienna, \textcode{sebastian.tschiatschek@univie.ac.at}\\
  \textsuperscript{3}Microsoft Research, 
  \textcode{sam.devlin@microsoft.com}
}

\begin{document}
\maketitle

\begin{abstract}
\looseness-1Block-based programming environments are increasingly used to introduce computing concepts to beginners. However, novice students often struggle in these environments, given the conceptual and open-ended nature of programming tasks. To effectively support a student struggling to solve a given task, it is important to provide adaptive scaffolding that guides the student towards a solution.
We introduce a scaffolding framework based on pop quizzes presented as multi-choice programming tasks. To automatically generate these pop quizzes, we propose a novel algorithm, \algmultihop. More formally, given a reference task with a solution code and the student's current attempt, \algmultihop~synthesizes new tasks for pop quizzes with the following features: (a) \ind~(i.e., individualized to the student's current attempt), (b) \interp~(i.e., easy to comprehend and solve), and (c) \conceal~(i.e., do not reveal the solution code). Our algorithm synthesizes these tasks using techniques based on symbolic reasoning and graph-based code representations.
We show that our algorithm can generate hundreds of pop quizzes for different student attempts on reference tasks from \emph{Hour of Code: Maze Challenge}~\cite{hourofcode_maze} and \emph{Karel}~\cite{intro_to_karel_codehs}. We assess the quality of these pop quizzes through expert ratings using an evaluation rubric. Further, we have built an online platform for practicing block-based programming tasks empowered via pop quiz based feedback, and report results from an initial user study.
%
\end{abstract}


\section{Introduction}\label{sec:intro}
\setcounter{footnote}{0}
\looseness-1The emergence of block-based visual programming platforms has made coding more interactive and appealing for novice students. Block-based programming uses ``code blocks'' that reduce the burden of syntax and focuses on key programming concepts. Led by the success of languages like \emph{Scratch}~\cite{DBLP:journals/cacm/ResnickMMREBMRSSK09}, initiatives like \emph{Hour of Code} by Code.org~\cite{hourofcode}, and online courses like \emph{Intro to Programming with Karel} by CodeHS.com \cite{intro_to_karel_codehs,pattis1981karel}, block-based programming has become integral to introductory CS education. 

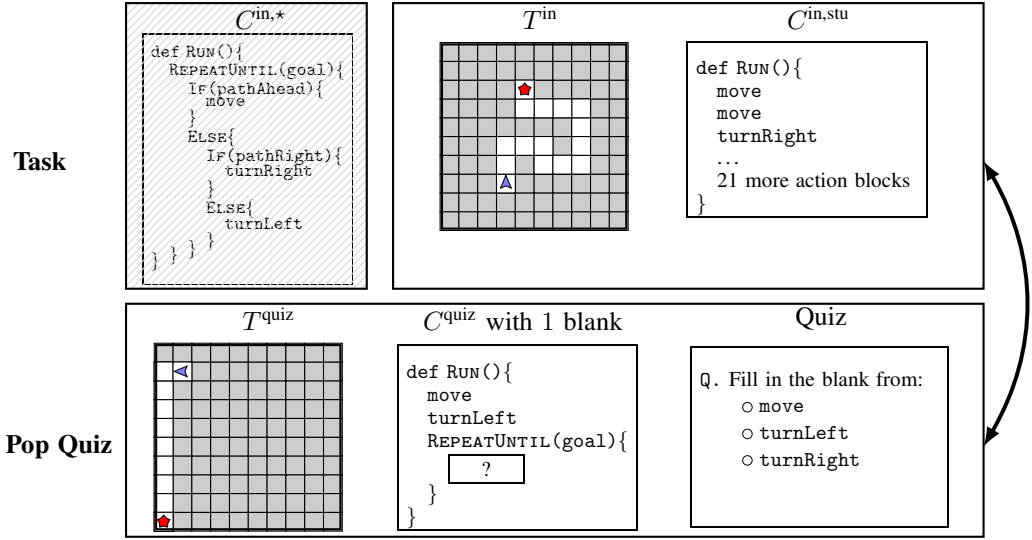
\begin{figure*}[!t]
\centering
\begin{tikzpicture}[
        box/.style={rectangle,draw=black, minimum size=0.25cm},
        ]
        \foreach \x in {-9.2,-8.95,...,-6.95}{
            \foreach \y in {2,2.25,...,4.25}
                \node[box, fill=gray!40] at (\x,\y){};
        }
       
        \foreach \y in {2.5,2.75,3}{
                \node[box, fill=white] at (-8.45,\y){};
        }
        \node[box, fill=white] at (-7.95,2.75){};
        \node[box, fill=white] at (-7.7,2.75){};
          \foreach \x in {-8.45,-8.2,-7.95}{
                \node[box, fill=white] at (\x,3){};
        }
        \foreach \y in {2.75,3,3.25, 3.5}{
                \node[box, fill=white] at (-7.45,\y){};
        }
        \foreach \x in {-7.45,-7.7,-7.95,-8.2}{
                \node[box, fill=white] at (\x,3.5){};
        }
        \node[box, fill=white] at (-8.2,3.75){};
        \draw[draw=black, thick] (-9.3,1.9) rectangle (-6.85,4.35);
        \node[draw, fill=red, star, star points=5,inner sep=0pt,minimum size=5pt] at (-8.2,3.75){};
        \node[draw, fill=blue!50, dart, rotate=90, inner sep=0.2pt,minimum size=4pt] at (-8.45,2.5){};
        \node[text width=0.5cm, anchor=west, right] at (-6.45, 3.5)
    {\begin{boxcode}{3.68cm}{0.80}{0.975}
				\textcode{def }\DSLRun\textcode{()\{}\\
                \quad \DSLMove\\
                \quad \DSLMove\\
                \quad \DSLTurnRight\\
                \quad \ldots\\
                \quad \text{21 more action blocks}\\
				\textcode{\}}
			\end{boxcode}};
	\draw[draw=black, thick] (-9.95,1.1) rectangle (-2.1, 4.9);
	\node[text width=0.5cm, anchor=east, left] at (-12.9, 3.1)
    {\begin{boxcode}{3.68cm}{0.70}{0.55}
				\textcode{def }\DSLRun\textcode{()\{}\\
				\quad \DSLRepeatUntil\textcode{(}\DSLBoolGoal\textcode{)\{}\\
                \quad \quad
              \DSLIf\textcode{(}\DSLBoolPathAhead\textcode{)\{}\\
				\quad \quad \quad \DSLMove\\ 
				\quad \quad \textcode{\}}\\
				\quad \quad 
				\DSLElse\textcode{\{}\\
				\quad \quad \quad \DSLIf\textcode{(}\DSLBoolPathRight\textcode{)\{}\\
				\quad \quad \quad \quad \DSLTurnRight\\
				\quad \quad \quad \textcode{\}}\\
				\quad \quad \quad
				\DSLElse\textcode{\{}\\
				\quad \quad \quad \quad \DSLTurnLeft\\
				\quad \quad \quad \textcode{\}}\\
			     \vspace{-3.5mm}
				\quad \quad \textcode{\}}\\
		     	\vspace{-4.5mm}				
				\quad \textcode{\}}\\
				\vspace{-5.5mm}				
				\textcode{\}}
			\end{boxcode}
			};
	\draw[draw=black, thick, pattern=north east lines, pattern color=gray!30] (-13.5,1.1) rectangle (-10.3, 4.9);
	\node[text width=0.5cm] at (-7.975,4.7) {$T^\text{in}$};
	\node[text width=0.5cm] at (-4.46,4.7) {\studentcode};
	\node[text width=0.5cm] at (-11.8,4.7) {\solutioncode};
	\draw[draw=black, thick] (-13.5,-2.2) rectangle (-2.1, 0.9);
	[
        box/.style={rectangle,draw=black, minimum size=0.25cm},
        ]
    
        \foreach \x in {-13,-12.75,...,-10.75}{
            \foreach \y in {-2,-1.75,...,0.25}
                \node[box, fill=gray!40] at (\x,\y){};
            
        }
    	
     \foreach \y in {-2,-1.75,...,-0.25, 0}{
    		\node[box, fill=white] at (-13,\y){};
    		
    	}
    	
    \foreach \x in {-13,-12.75}{
    	\node[box, fill=white] at (\x,0){};
    		
    }

        \node[box, fill=white] at (-13,-2){};
        \draw[draw=black, thick] (-13.1,-2.1) rectangle (-10.65,0.35);
         \node[draw, fill=red, star, star points=5,inner sep=0pt,minimum size=5pt] at (-13,-2){};
         \node[draw, fill=blue!50, dart, rotate=180, inner sep=0.2pt,minimum size=4pt] at (-12.75,0){};
    \node[text width=0.5cm, anchor=west, right] at (-10.3, -0.6)
    {\begin{boxcode}{3.68cm}{0.80}{1}
				\textcode{def }\DSLRun\textcode{()\{}\\
				\quad \DSLMove\\
			   \quad \DSLTurnLeft\\
			   \quad
		 \DSLRepeatUntil\textcode{(}\DSLBoolGoal\textcode{)\{}\\
				\quad \quad \framebox[8.0\width]{?}\\
				\quad \textcode{\}}\\
				\textcode{\}}
			\end{boxcode}};
	 \node[text width=0.5cm, anchor=west, right] at (-6.4, -0.6)
	 {\begin{boxcode}{3.98cm}{0.8}{1.15}
				\textcode{Q.} \text{Fill in the blank from: }\\
				\quad \quad \tikz\draw[black,fill=none] (0,0) circle (.5ex); \DSLMove\\
				\quad \quad \tikz\draw[black,fill=none] (0,0) circle (.5ex); \DSLTurnLeft\\
				\quad \quad \tikz\draw[black,fill=none] (0,0) circle (.5ex); \DSLTurnRight\\
				\\
			\end{boxcode}
	 };
	\node[text width=0.5cm] at (-11.7,0.7) {$T^\text{quiz}$};
	\node[text width=2.6cm] at (-8.25,0.7) {$C^\text{quiz}\text{ with }1\text{ blank}$};
	\node[text width=1cm] at (-4.1,0.7) {Quiz};
	\tikzstyle{doublearr}=[latex-latex, black, line width=1.5pt]
    \draw [doublearr, bend left]    (-2.1, 2.8) to (-2.1,-1);
    \node[text width=1cm] at (-14.5,2.8){\textbf{Task}};
    \node[text width=1.7cm] at (-14.25,-1){\textbf{Pop Quiz}};
    \end{tikzpicture}

\caption{Illustration of our pop quiz based framework. The ``Task'' panel shows an input task $T^\text{in}$ from HOC~\cite{hourofcode_maze}, the student's current attempt \studentcode, and the solution code \solutioncode~(not revealed to the student). The student is currently unsuccessful in solving the task: the current attempt \studentcode~does not solve the visual puzzle within the maximal number of permitted blocks ($7$ blocks) and does not use any of the required constructs  (\DSLRepeatUntil and \DSLIfElse constructs). The ``Pop Quiz'' panel shows a pop quiz generated by our algorithm in the form of task-code pair ($T^\text{quiz}, C^\text{quiz}$) along with a multiple choice question, introducing the \DSLRepeatUntil~construct. After the student solves the pop quiz, they resume working on the input task. The framework would be invoked when a student needs help; importantly, the pop quizzes presented to the student are adaptive w.r.t. the student's current attempt \studentcode. Moreover, our algorithm generates pop quizzes that are easy to comprehend and solve, and $C^\text{quiz}$ sufficiently conceals \solutioncode.
}
\label{fig:intro}
\end{figure*}

\looseness-1Programming tasks on these platforms are conceptual and open-ended, requiring multi-step deductive reasoning to solve, thereby making them challenging for students. To effectively support a struggling student to solve a particular task, it is important to provide feedback on their attempts. However, on platforms that have millions of students, it is infeasible for human tutors to provide feedback. Hence, there is a critical need for automated feedback generation systems to provide personalized support to  students~\cite{adaptivefeedbackmarwan20,DBLP:journals/corr/abs-2102-05741}. Existing work in the domain has explored various methods of personalized feedback generation within a task, such as providing next-step hints in the form of next code blocks to use in a student attempt~\cite{DBLP:conf/its/RiversK14,DBLP:conf/lats/PiechSHG15,DBLP:journals/aiedu/RiversK17,DBLP:conf/kbse/ZimmermanR15,DBLP:conf/edm/PriceZB17}, providing adaptive worked examples~\cite{DBLP:conf/aied/PriceZB17,DBLP:conf/edm/ZhiMDLPB19,DBLP:conf/sigcse/PriceDL17}, and providing data-driven analysis of a student's misconceptions~\cite{DBLP:conf/sigcse/CherenkovaZP14,DBLP:conf/iwpc/WieseRKA19,DBLP:conf/icse/WieseRF19,DBLP:conf/aaai/WuMGP19,DBLP:journals/corr/abs-1708-06564,DBLP:conf/edm/EfremovGS20,DBLP:conf/iclr/BunelHDSK18}. 

\looseness-1In this paper, we investigate an alternate method of personalized feedback generation that guides a student towards a task's solution while involving inquiry-driven and problem-solving aspects~\cite{DBLP:conf/sigcse/CordovaCGW21}. In particular, we introduce a scaffolding framework based on pop quizzes that contain new programming tasks presented as multi-choice questions.\footnote{We refer to these multi-choice questions as ``pop quizzes'' as the framework could present these quizzes whenever a student needs help~\cite{DBLP:conf/sigcse/Cicirello09}.}
Our framework is inspired by prior studies that showed the efficacy of multi-choice questions in helping novice students learn to code \cite{DBLP:conf/chi/PriceWSM20,DBLP:journals/ijet/ZhangLZH20,DBLP:conf/sigcse/Grover21,DBLP:conf/iticse/SoltanpoorTD18,DBLP:conf/ecai2/EneS19}. The framework is designed to be invoked as follows: Given a task and a student's current unsuccessful attempt, the framework can help the student by presenting a pop quiz intended to resolve their misconception. For the scaffolding to be effective, we center the design of the new programming task for a pop quiz around three features: \ind, \interp, and \conceal; see details in Fig.~\ref{fig:intro} and Section~\ref{sec:notation.setup}.
However, hand-crafting these new quizzes is time-consuming and potentially error-prone when required for a large number of tasks and different student attempts.
To this end, we seek to \emph{automatically} generate these pop quizzes by synthesizing new programming tasks.

\subsection{Key Challenges and Our Contributions}\label{sec:intro.contributions}

\looseness-1There are several challenges in synthesizing new visual programming tasks with the above mentioned features, including the following:
(i) current techniques for synthesizing visual programming tasks do not adapt to student attempts~\cite{DBLP:conf/nips/AhmedCEFGRS20}; (ii) the mapping from the space of visual tasks to their solution codes is highly discontinuous as shown in \cite{DBLP:conf/nips/AhmedCEFGRS20}, and hence task mutation based techniques are ineffective~\cite{DBLP:conf/aaai/SinghGR12,DBLP:conf/ijcai/PolozovOSZGP15}; (iii) the space of possible tasks and their solutions is potentially unbounded, and hence techniques that rely on exhaustive enumeration are intractable~\cite{DBLP:conf/aaai/SinghGR12,DBLP:conf/ijcai/AhmedGK13,DBLP:conf/aaai/AlvinGMM14}.

\looseness-1In this work, we develop a novel algorithm, \algmultihop, that synthesizes pop quizzes with the desirable features of our scaffolding framework.
Our algorithm overcomes the above-mentioned challenges by using techniques of symbolic execution, search algorithms, and graph-based code representations. Our key contributions are: \textbf{(I)} We present a modular and extensible algorithm for generating pop quizzes that operates in three stages (see Sections~\ref{sec:notation}~and~\ref{sec:model});\footnote{Implementation of the algorithm is publicly available at\\ \text{\quad \ \ \  }\url{https://github.com/machine-teaching-group/aied2022\_pquizsyn\_code}} \textbf{(II)} We show that our approach can generate hundreds of pop quizzes for different types of student attempts on reference tasks from real-world programming platforms (see Section \ref{sec:evaluation}); \textbf{(III)} We assess the quality of our algorithm through expert ratings using a multi-dimensional evaluation rubric (see Section~\ref{sec:surveystudy}); \textbf{(IV)} We have built an online platform with our framework and demonstrate the utility of pop quiz based feedback through an initial user study (see Section~\ref{sec:userstudy}).\footnote{Online platform is publicly available at \url{https://www.teaching-blocks-hints.cc/}\label{footnote:webapp}}
%

\subsection{Additional Related Work}
\looseness-1\textbf{Feedback via modelling programming concepts.} Apart from the above-mentioned methods such as next-step hints, there has been extensive work on feedback generation via modelling programming concepts. Here, several techniques have been proposed, including: (a) detecting challenging concepts by analyzing student attempts~\cite{DBLP:conf/sigcse/CherenkovaZP14,DBLP:conf/iwpc/WieseRKA19,DBLP:conf/icse/WieseRF19}; (b) discovering student misconceptions using task-specific rubrics and neural program embeddings~\cite{DBLP:conf/aaai/WuMGP19}; (c) defining concepts through knowledge components~\cite{DBLP:conf/icer/RiversHK16,DBLP:conf/sigcse/CrichtonSH21,DBLP:conf/edm/AkramAMWMNBL20}.
%

\looseness-1\textbf{Evaluation of feedback methods.} An important aspect to consider when developing feedback generation methods is their evaluation criteria. Most next-step feedback generation methods are evaluated based on expert annotations or automated procedures ~\cite{DBLP:conf/lats/PiechSHG15,DBLP:journals/corr/abs-1708-06564,DBLP:journals/aiedu/PriceDZPLCB19}. In contrast, example-driven feedback techniques are typically evaluated using a multi-dimensional rubric~\cite{DBLP:conf/aied/PriceZB17,DBLP:conf/edm/ZhiMDLPB19}. In our work, we evaluate the scaffolding framework through expert ratings using a rubric, as well as an initial user study. 
%
\section{Problem Setup and Definitions}\label{sec:notation}
In this section, we formalize our objective and introduce important technical definitions.

\subsection{Problem Setup}\label{sec:notation.setup}

\looseness-1\textbf{Task space.} We define the space of tasks as~\taskspace. A task $T \in \taskspace$ consists of a visual puzzle and a set of available types of code blocks (e.g.,~\DSLMove,~\DSLRepeatUntil) allowed in the solution code.
Additionally, the solution code must be within a certain size threshold in terms of the number of code blocks. We denote the current task that a student is solving as $T^\text{in} \in \taskspace$; see $T^\text{in}$ in Fig.~\ref{fig:intro}. In this work, we use tasks from \emph{Hour of Code: Maze Challenge}~\cite{hourofcode_maze} by Code.org~\cite{codeorg} and \emph{Intro to Programming with Karel}~\cite{intro_to_karel_codehs} by CodeHS.com~\cite{codehscom}; henceforth, we refer to them as HOC and Karel tasks, respectively.

\looseness-1\textbf{Code space.} We define the space of all possible codes as \codespace~and represent them using a \emph{Domain Specific Language }(DSL)~\cite{gulwani2017program}. In particular, for codes relevant for HOC and Karel tasks, we use a DSL based on~\cite{DBLP:conf/nips/AhmedCEFGRS20}.
A code $C \in \codespace$ has the following attributes: $C_\text{blocks}$ is the set of types of code blocks used in $C$, $C_\text{size}$ is the number of blocks used, and $C_\text{depth}$ is the depth of the \emph{Abstract Syntax Tree} of $C$. We denote a distance metric in this space as $D_\codespace$. For a given $C \in \codespace$ and a positive integer $l$, we define a neighborhood function as $\mathcal{N}_\codespace(C,l) = \{ C'~| ~D_\codespace(C', C) \leq l\}$. The solution code $\solutioncode \in \codespace$ for the task $T^\text{in}$ solves the visual puzzle using the allowed types of code blocks within the specified size threshold. A student attempt for $T^\text{in}$ is denoted as~$\studentcode \in \codespace$.

\looseness-1\textbf{Objective.} For an input task $T^\text{in}$ with solution code \solutioncode~and given the current student attempt \studentcode, our objective is to generate a pop quiz in form of a new task-code pair $(T^\text{quiz}, C^\text{quiz})$ designed on the basis of the following features: (i) \emph{Adaptive}, i.e., $C^\text{quiz}$ accounts for \solutioncode~and \studentcode, ensuring that $C^\text{quiz}$ is individualized to the student's current attempt; (ii) \emph{Comprehensible}, i.e., $C^\text{quiz}$ solves $T^\text{quiz}$ correctly and the pop quiz is easy to comprehend/solve without confusing the student; (iii) \emph{Concealing}, i.e., $D_\codespace(C^\text{quiz},\solutioncode)$ is high, ensuring that $C^\text{quiz}$ sufficiently conceals the solution code \solutioncode{} and does not directly reveal it in order to encourage problem-solving aspects.

\subsection{Technical Definitions}
 \label{sec:notation.definitions}

\looseness-1\textbf{Sketch space.} We capture the key conceptual elements of a code using a higher level abstraction called a \textit{sketch}~\cite{DBLP:conf/aaai/SinghGR12,DBLP:conf/ijcai/AhmedGK13}. The sketch of a code preserves its important programming constructs. Similar to the code DSL, we define the sketch space~\sketchspace{} using a sketch DSL based on~\cite{DBLP:conf/nips/AhmedCEFGRS20}. 
Similar to the \emph{Abstract Syntax Tree} representation of a code, we represent a sketch as a tree having the programming constructs as its nodes. The mapping from the code space to the sketch space is captured by the many-to-one map, $\sketchmap\colon \codespace \rightarrow \sketchspace$, i.e., the representation of a code $C$ in \sketchspace~is given by \sketchmap($C$). As \sketchspace~is an abstraction of \codespace, multiple elements of \codespace~can correspond to a single element in~\sketchspace. Similar to $D_\codespace$ and $\mathcal{N}_\codespace$, we denote a distance metric in the sketch space as $D_\sketchspace$ and a neighborhood function as $\mathcal{N}_\sketchspace(S,l) = \{ S'~|~D_\sketchspace(S', S) \leq l\}$ for a given $S \in \sketchspace$ and a positive integer $l$.

\looseness-1\textbf{Sketch substructures.} For a sketch $S$, we define a substructure as a sub-tree containing the nodes of $S$ up to a particular depth and sharing the same root node; note that a substructure of a sketch is also a sketch. We denote the set of all substructures of $S$ as $\substructures(S) \subseteq \sketchspace$; the size of the set $\substructures(S)$ is typically small.
For example, the sketch shown in Fig.~\ref{fig:pipeline.solsketch} has the following $4$ substructures: (i) \begin{small}{\{\DSLRun{}\}}\end{small}, (ii) \begin{small}{\{\DSLRun\{\DSLRepeatUntil{}(\DSLBoolGoal){}\}\}}\end{small}, (iii) \begin{small}{\{\DSLRun\{\DSLRepeatUntil{}(\DSLBoolGoal)\{\DSLIfElse(\SDSLBool){}\}\}\}}\end{small}, and (iv) \begin{small}{\{\DSLRun\{\DSLRepeatUntil{}(\DSLBoolGoal)\{\DSLIfElse(\SDSLBool)\{\{\}; \{\DSLIfElse(\SDSLBool){}\}\}\}\}\}}\end{small}.

\looseness-1\textbf{Code reductions.} For a code $C \in \codespace$ with sketch $S := \sketchmap(C)$, consider one of the sketches $S_\text{sub} \in \substructures(S)$. We define the set of code reductions of $C$ w.r.t. sketch $S_\text{sub}$ as all codes obtained by removing one or more nodes of $C$ while preserving the sketch $S_\text{sub}$; note that the reduction of a code is also a code.  We denote the set of all reductions as $\reducedcodes(C~|~S_\text{sub}) \subseteq \codespace$. 
For example, for \solutioncode~in Fig.~\ref{fig:intro} and \begin{small}$S_\text{sub} = \{\DSLRun\{\DSLRepeatUntil($\DSLBoolGoal{}$)\}\}$\end{small}, the set  $\reducedcodes(\solutioncode|~S_\text{sub})$ has the following $3$ codes: (i) \begin{small}{\{\DSLRun\{\DSLRepeatUntil{}(\DSLBoolGoal)\{\DSLMove{}\}\}\}}\end{small}, (ii) \begin{small}{\{\DSLRun{}\{\DSLRepeatUntil{}(\DSLBoolGoal)\{\DSLTurnRight{}\}\}\}}\end{small}, and (iii) \begin{small}\{\DSLRun{}\{\DSLRepeatUntil{}(\DSLBoolGoal)\{\DSLTurnLeft{}\}\}\}\end{small}.

\section{Our Algorithm \algmultihop} \label{sec:model}
In this section, we present our algorithm that generates pop quizzes via synthesizing new tasks. One might be tempted to synthesize tasks by first generating a new visual puzzle and then obtaining its solution code. As discussed in Section~\ref{sec:intro} and shown in~\cite{DBLP:conf/nips/AhmedCEFGRS20}, the mapping from the space of visual tasks to their solution codes is highly discontinuous and reasoning about desirable tasks directly in the task space is ineffective. 
However, the task synthesis algorithm from~\cite{DBLP:conf/nips/AhmedCEFGRS20} is not applicable to our work as we seek to generate tasks that also account for the student's current attempt. 
To this end, we develop a novel algorithm \algmultihop~(\emph{Programming Pop Quizzes via Synthesis}) that generates tasks adaptive to the student's current attempt. Our algorithm operates in three stages: (i) Stage 1 generates a sketch based on the task's solution code and the student's current attempt; (ii) Stage 2 instantiates this sketch in the form of a new task-code pair; (iii) Stage 3 generates the pop quiz from the new task-code pair. Fig.~\ref{fig:pipeline.abstract} illustrates these stages, and details are provided below. 
%

\begin{figure*}[!t]
\centering
\begin{minipage}{0.65\textwidth} 
\begin{minipage}{1\textwidth}
    \begin{subfigure}[b]{1.0\textwidth}
    	\centering
        \begin{tikzpicture}[scale=0.75, every node/.style={transform shape}]
             \draw (0,0) rectangle (0.8,0.8) node[pos=0.5] {\small$\studentcode$};
             \draw (0,1.5) rectangle (0.8,2.3) node[pos=0.5] {$\solutioncode$};
             \draw (-1.2,0.6) rectangle (-0.4,1.4) node[pos=0.5] {$T^\text{in}$};
             \draw[pattern=dots, pattern color=gray!40] (1.3,-0.2) rectangle (5.8,2.5) node[pos=0.5] {};
             \draw[fill=white] (1.5,0) rectangle (2.3,0.8) node[pos=0.5] {$S^\text{in,stu}$};
             \draw[fill=white] (1.5,1.5) rectangle (2.3,2.3) node[pos=0.5] {$S^{\text{in,}\star}$};
             \node [trapezium, trapezium angle=85, minimum width=15mm, fill=white, draw] at (3.25,1.2){${\hintpolicy}$};
             \draw[fill=white] (4.8,0.85) rectangle (5.6,1.65) node[pos=0.5] {$S^\text{quiz}$};
             \node[] at (5.3,2.25) {\textbf{\large\sketchspace}};
             \draw (3.1, -0.1) rectangle (4.2, 0.4) node[pos=0.5] {Stage 1};
             \draw (6.05, -0.1) rectangle (7.45, 0.4) node[pos=0.5] {Stage 2(i)};
             \draw (7.65, -0.1) rectangle (9.15, 0.4) node[pos=0.5] {Stage 2(ii)};
             \draw (6.3, 2.2) rectangle (7.7, 2.6) node[pos=0.5] {Stage 3};
             \draw (6.5,0.85) rectangle (7.3,1.65) node[pos=0.5] {$C^\text{quiz}$};
             \draw (8,0.85) rectangle (8.8,1.65) node[pos=0.5] {$T^\text{quiz}$};
             \draw (8,1.85) rectangle (8.8,2.65) node[pos=0.5] {\small$C^{\text{\tiny{quiz,}}k}$};
             \draw[thick,->] (0.8,0.4) to node[pos=0.5, above]{$\sketchmap$} (1.3,0.4);
             \draw[thick,->] (0.8,1.9) to node[pos=0.5, above]{$\sketchmap$} (1.3,1.9);
             \draw[thick,->] (5.8,1.25) to node[pos=0.5, above]{$\sketchmap^{-1}$} (6.5,1.25);
             \draw[thick,->] (2.3,0.5) to node[pos=0.5, above]{} (2.8,0.95); 
             \draw[thick,->] (2.3,2) to node[pos=0.5, above]{} (2.9,1.5);
             \draw[thick,->] (4.4,1.25) to node[pos=0.5, above]{} (4.8,1.25);
             \draw[thick,->] (7.3,1.25) to node[pos=0.5, above]{} (8,1.25);
        	\draw[thick,->] (7,1.65) to node[pos=0.25, above]{\exposureparam} (8,2.25);
        \end{tikzpicture}
        \caption{Our algorithm \algmultihop}
        \label{fig:pipeline.abstract}
        \vspace{2mm}
    \end{subfigure}
\end{minipage}
\begin{minipage}{1\textwidth}
    \begin{minipage}{0.31\textwidth}
        \begin{subfigure}[b]{1\textwidth}
        	\centering
            \begin{boxcode}{3.62cm}{0.70}{0.95}
                \textcode{def }\DSLRun\textcode{()\{}\\
        		\ \ \DSLRepeatUntil\textcode{(}\DSLBoolGoal\textcode{)\{}\\
                \quad \quad
                \DSLIf\textcode{(}\SDSLBool\textcode{)\{\}}\\
        		\quad \quad 
        		\DSLElse\textcode{\{}\\
        		\quad \quad \quad \DSLIf\textcode{(}\SDSLBool\textcode{)\{\}}\\
        		\quad \quad \quad
        		\DSLElse\textcode{\{\}}\\
        		\quad \quad \textcode{\}}\\
        		\ \  \textcode{\}}\\
        		\textcode{\}}
            \end{boxcode}
            \vspace{-3.3mm}
            \caption{$S^{\text{in,}\star}$}    
            \label{fig:pipeline.solsketch}
        \end{subfigure}
    \end{minipage}
    \begin{minipage}{0.33\textwidth}
        \begin{subfigure}[b]{1\textwidth}
        	\centering
            \begin{boxcode}{3.68cm}{0.70}{1.0}
        		\textcode{def }\DSLRun\textcode{()\{\}}
            \end{boxcode}
            \vspace{-3mm}
            \caption{$S^{\text{in,stu}}$}
            \vspace{3mm}
            \label{fig:pipeline.stusketch}
        \end{subfigure}
        \begin{subfigure}[b]{1\textwidth}
        	\centering
        	\begin{boxcode}{3.68cm}{0.70}{1.0}
        		\textcode{def }\DSLRun\textcode{()\{}\\
        		\ \ \DSLRepeatUntil\textcode{(}\DSLBoolGoal\textcode{)\{\}}\\
        		\textcode{\}}
            \end{boxcode}
            \vspace{-3mm}
            \caption{$S^{\text{quiz}}$}
            \label{fig:pipeline.sketchhint}    
        \end{subfigure}
    \end{minipage}
    \begin{minipage}{0.32\textwidth}
        \begin{subfigure}[b]{1\textwidth}
        	\centering
            \begin{boxcode}{3.68cm}{0.70}{1.0}
        		\textcode{def }\DSLRun\textcode{()\{}\\
        		\quad \DSLMove\\
        		\quad \DSLTurnLeft\\
        		\quad \DSLRepeatUntil\textcode{(}\DSLBoolGoal\textcode{)\{}\\
        		\quad \quad \DSLMove\\
        		\quad \textcode{\}}\\
        		\textcode{\}}
        		\vspace{5.4mm}
            \end{boxcode}
            \vspace{-3mm}
            \caption{$C^{\text{quiz}}$}
            \label{fig:pipeline.codequiz}
        \end{subfigure}
    \end{minipage}
\end{minipage}
\vspace{-3mm}
\caption{ (a) illustrates \algmultihop. In particular, we can instantiate the presented algorithm using input task~$T^\text{in}$, its solution code \solutioncode, and the current student attempt \studentcode~from Fig.~\ref{fig:intro}. The sketch of \solutioncode~is shown in (b), sketch of \studentcode~is shown in (c), sketch of $C^\text{quiz}$ is shown in (d), and the code of the pop quiz $C^\text{quiz}$ is shown in (e).}
%
\label{fig:pipeline}
\end{minipage}
\hspace{2mm}
\begin{minipage}{0.32\textwidth}
    \begin{subfigure}[b]{1.0\textwidth}
    	\centering
        	\begin{tikzpicture}[scale=0.75, every node/.style={transform shape}]
                \draw[pattern=dots, pattern color=gray!40] (-0.1,0) rectangle (5.4,7.5) node[pos=0.5] {};
                \node[] at (5,6.5) {\textbf{\large\sketchspace}};
                \node[circle, fill=black, minimum size=0.15cm, inner sep=0pt, outer sep=0pt, label=right:{$S^{\text{in,stu}}$}] (nstu) at (2.75,4.7) {};
                \node [outer sep=1pt,fill=white,above] at (2.85,0.4) {\footnotesize{$S^{\text{in,}\star}$}};
                \draw (2.75,1.1) node[cross=4] {};
                \draw (2.75,4.5) ellipse (1.5cm and 1.5cm); 
                \draw (2.75,4.2) ellipse (2cm and 2.5cm); 
                \draw (2.75,3.8) ellipse (2.4cm and 3.6cm); 
                \draw (3,3.9) node[cross=4] {}; 
                \node [outer sep=1pt,fill=white,above] at (2.8,3.2) {\scriptsize{\{\DSLRun\{\DSLRepeatUntil{(\DSLBoolGoal)}}\}\}};
                \draw (3.2,2.6) node[cross=4] {}; 
                 \node [outer sep=1pt,fill=white,above] at (2.9,1.8) {\scriptsize{\{\DSLRun\{\DSLRepeatUntil(\DSLBoolGoal)\{\DSLIfElse(\SDSLBool){}\}}\}\}};
                \node [outer sep=1pt,fill=white,above] at (2.75,5.25) {$l=1$};
                 \node [outer sep=1pt,fill=white,above] at (2.75,6) {$l=2$};
                  \node [outer sep=1pt,fill=white,above] at (2.75,6.8) {$l=3$};
            \end{tikzpicture}
    \end{subfigure}
\vspace{-4mm}    
\caption{
\algmultihop{} Stage~1 for the scenario shown in Fig.~\ref{fig:intro}.  $\textnormal{X}$ shows substructures of $S^{\text{in,}\star}$ in $l$-hop neighborhoods of $S^\text{in,stu}$ for $l \in \{1, 2, 3\}$. Details are provided in Section~\ref{sec:model.stage1}.
}
\label{fig.approach.algmultihop}
\end{minipage}
\end{figure*}

\subsection{Stage 1: Generating the Pop Quiz Sketch $S^\text{quiz}$}\label{sec:model.stage1}

\looseness-1We begin by describing Stage 1 of our algorithm as illustrated in Fig.~\ref{fig:pipeline.abstract}. In this stage, \hintpolicy{} routine returns a suitable sketch $S^\text{quiz}$ that is instantiated in the later stages. The input to the routine is the student sketch $S^{\text{in,stu}} := \sketchmap(\studentcode)$ and solution sketch $S^{\text{in,}\star} := \sketchmap(\solutioncode)$. By operating on the sketch space first, we can generate meaningful and adaptive codes in the later stages.
To generate pop quizzes based on the features mentioned in Section~\ref{sec:notation.setup}, we require the sketch of the pop quiz $S^\text{quiz}$ to have the following attributes: (i) $S^\text{quiz}$ should direct the student towards the solution sketch $S^{\text{in,}\star}$, i.e., $D_\sketchspace(S^\text{quiz}, S^{\text{in,}\star})$ should be low; (ii) $S^\text{quiz}$ should be adaptive w.r.t. the student's sketch $S^\text{in,stu}$, i.e., $S^\text{quiz} \in \mathcal{N}_\sketchspace(S^\text{in,stu},l)$ for a low value of $l$. While these conditions ensure that $S^\text{quiz}$ directs the student towards the solution sketch and is adaptive, it could potentially lead to a sketch that does not belong to the set of substructures of the solution sketch, i.e., $S^\text{quiz} \notin \substructures(S^{\text{in,}\star})$---in that case, there is no valid code reduction of \solutioncode~w.r.t. $S^\text{quiz}$ (see Section~\ref{sec:notation.definitions}) and this makes it challenging to instantiate sketches into desirable codes $C^\text{quiz}$ (see algorithm variant \alghintpolicystruct~in Section~\ref{sec:surveystudy} and Footnote~\ref{footnote:sec3}).
Hence, \hintpolicy~generates $S^\text{quiz}$ as follows (see Fig.~\ref{fig.approach.algmultihop}):
\begin{enumerate}[(i)]
    \item Pick $\hat{l}$ as $\min l \in \{1,2,\ldots\}$ s.t. $\mathcal{N}_\sketchspace(S^\text{in,stu},~l) \cap \substructures(S^{\text{in,}\star})$ is non-empty.
    \item Generate $S^\text{quiz} \in \argmin_{S \in \mathcal{N}_\sketchspace(S^\text{in,stu}, ~\hat{l}) ~\cap~ \substructures(S^{\text{in,}\star})} D_\sketchspace(S, S^{\text{in,}\star})$.    
\end{enumerate}

\subsection{Stage 2: Synthesizing $(T^\text{quiz}, C^\text{quiz})$ from $S^\text{quiz}$}\label{sec:model.stage2}

\looseness-1Next, we describe Stage 2 of our algorithm. We first generate $C^\text{quiz}$ from $S^\text{quiz}$, as illustrated in Stage 2(i) of Fig.~\ref{fig:pipeline.abstract}. Specifically, for a sketch $S^\text{quiz}$ generated in Stage 1, 
we employ the \emph{code mutation} methodology proposed in~\cite{DBLP:conf/nips/AhmedCEFGRS20} to obtain a code $C^\text{quiz}$. However, this methodology requires a meaningful starting code $C^\text{seed}$. Since $S^\text{quiz} \in \substructures(S^{\text{in,}\star})$ by the design of Stage 1, we begin by picking $C^\text{seed}$ from the set \reducedcodes($\solutioncode|~S^\text{quiz}$).\footnote{When $S^\text{quiz} \notin \substructures(S^{\text{in,}\star})$, we set $C^\text{seed}$ as a random instantiation of $S^\text{quiz}$ -- see algorithm variant \alghintpolicystruct~in Section~\ref{sec:surveystudy}.\label{footnote:sec3}}
The methodology of~\cite{DBLP:conf/nips/AhmedCEFGRS20} provides us multiple code mutations of $C^\text{seed}$. The extent to which these code mutations differ from $C^\text{seed}$ and \solutioncode~is controlled by the constraints imposed based on the values of the boolean variables, conditionals, and action blocks (\begin{small}\DSLMove, \DSLTurnLeft, \DSLTurnRight, \DSLPickMarker, \DSLPutMarker{}\end{small}) of $C^\text{seed}$, as well as constraints on the size of the obtained code. Specifically, these mutations allow us to control the extent to which $D_\codespace(C^\text{quiz},\solutioncode)$ varies, which is a desired feature as stated in Section ~\ref{sec:notation.setup}. 

\looseness-1Next, we generate a new task $T^\text{quiz}$ from a code $C^\text{quiz}$ as illustrated in Stage 2(ii) of Fig.~\ref{fig:pipeline.abstract}.
Specifically, we generate $T^\text{quiz}$ such that its solution code is $C^\text{quiz}$. We achieve this using techniques of symbolic execution and best-first search, building on the task synthesis methodology presented in~\cite{DBLP:conf/nips/AhmedCEFGRS20}. 
%

\subsection{Stage 3: Generating Multi-Choice Question from $(T^\text{quiz}, C^\text{quiz})$}
%
\looseness-1In this stage, we generate a pop quiz with a fixed set of answer choices; see Figs.~\ref{fig:intro}~and~\ref{fig:karel.illustration}. We pick a task-code pair $(T^\text{quiz}, C^\text{quiz})$, and expose only a part of $C^\text{quiz}$ determined by an exposure parameter $k$, i.e., $C^\text{quiz}$~contains $k$ blanks. These blanks must be filled out by the student from the set of answer choices in a manner that would solve $T^\text{quiz}$.
Specifically, we generate the pop quiz with $\exposureparam = 1$ blanks. To obtain the blank for the quiz, we do an in-order traversal of $C^\text{quiz}$ and leave out the last leaf node as blank.
%


\begin{figure}[!t]
\renewcommand{\arraystretch}{0.9}
\centering
	\scalebox{0.89}{\begin{tabular}{l|c|c|rr}
		    \multicolumn{1}{c|}{Name, source for ${T^{\text{in}}}$} &
		    \multicolumn{1}{c|}{
		    ${C^{\text{in,}\star}_\text{size}}$, ${S^{\text{in,}\star}}$ for ${T^{\text{in}}}$}
		    &
		    \multicolumn{1}{c|}{${S^\text{quiz}}$ \scriptsize$\in \substructures({S^{\text{in,}\star}})$\normalsize} &
		 \multicolumn{1}{c}{${\#C^\text{quiz}}$} & \multicolumn{1}{c}{${\#T^\text{quiz}}$}
			 \\
			\toprule
			\multirow{3}{*}{
			\shortstack{
			T-1
			\\
		    \footnotesize{\textbf{HOC:Maze08}~\cite{hourofcode_maze}}
			}
			}
			& 
			\multirow{3}{*}{
			\shortstack[c]{
			$6$ 
			\\
		   \text{\scriptsize{\{\DSLRun\{\DSLRepeat; \DSLRepeat{}\}\}}}
			}
			}
			& 
			\text{\scriptsize{\{\DSLRun{}}\}} 
			& 
           	$22$ & $220$
			\\
			&&
			\text{\scriptsize{\{\DSLRun\{\DSLRepeat{}}\}\}} 
			& $34$ & $340$
			\\
			&&
			\ $S^{\text{in,}\star}$
			& $179$ & $1790$
			\\
			\hline
			\multirow{3}{*}{\shortstack{
			T-2
			\\
			\footnotesize{\textbf{HOC:Maze16}~\cite{hourofcode_maze}}
			}}&
			\multirow{3}{*}{\shortstack[c]{
			$5$ 
			\\
		\text{\scriptsize{\{\DSLRun\{\DSLRUntil\{\DSLIf{}\}\}\}}}
			}
			}
			&
			\text{\scriptsize{\{\DSLRun{}}\}} 
			& 
			$10$ & $100$
			\\
            &&
            \text{\scriptsize{\{\DSLRun\{\DSLRUntil{}\}\}}} 
			& $6$ & $60$
			\\
			&& 
			\ $S^{\text{in,}\star}$
			& $19$ & $190$
			\\
			\hline
			\multirow{3}{*}{\shortstack{
			T-3
			\\
			\footnotesize{\textbf{HOC:Maze18}~\cite{hourofcode_maze}}
			}}&
			\multirow{3}{*}{\shortstack[c]{
		    $5$
			\\
		 \text{\scriptsize{\{\DSLRun\{\DSLRUntil\{\DSLIfElse{}\}\}}}
			}
			}
			& 
			\text{\scriptsize{\{\DSLRun{}\}}} 
			& 
			$10$ & $100$
			\\
			&&
			\text{\scriptsize{\{\DSLRun\{\DSLRUntil{}\}\}}} 
			& $6$ & $60$
			\\
			&&
			\ $S^{\text{in,}\star}$
			& $9$ & $90$
			\\
			\hline
			\multirow{4}{*}{\shortstack{
			T-4
			\\
		    \footnotesize{\textbf{HOC:Maze20}~\cite{hourofcode_maze}}
			}}
			&
			\multirow{4}{*}{
			\shortstack[c]{
			$7$ 
			\\
			\scriptsize{\{\DSLRun\{\DSLRUntil\{\DSLIfElse\{\{\};\{\DSLIfElse{}\}\}\}\}\}}
			}
			}
			&
			\text{\scriptsize{\{\DSLRun{}}\}} 
			& $10$ & $100$
			\\
			&&
			\text{\scriptsize{\{\DSLRun\{\DSLRUntil{}\}\}}} 
			& $6$ & $60$
			\\
			&&
			\text{\scriptsize{\{\DSLRun\{\DSLRUntil\{\DSLIfElse{}\}\}\}}} 
			& $9$ & $90$
			\\
			&& 
			\ $S^{\text{in,}\star}$
			& $10$ & $100$
			\\
			\hline
			\multirow{3}{*}{\shortstack{
			T-5
			\\
			\footnotesize{\textbf{Karel:Opposite}~\cite{intro_to_karel_codehs}}
			}}
			 &
			\multirow{3}{*}{
			\shortstack[c]{
			$6$
			\\
			\text{\scriptsize{\{\DSLRun\{\DSLRepeat\{\DSLIfElse{}\}\}\}}}
			}
			}
			&
			\text{\scriptsize{\{\DSLRun{}}\}} 
			& $73$ & $730$
			\\
			&&
			\text{\scriptsize{\{\DSLRun\{\DSLRepeat{}\}\}}} 
			& $118$ & $1180$
			\\
			&& 
			\ $S^{\text{in,}\star}$
			& $343$ & $3430$
			\\
            \hline
			\multirow{2}{*}{\shortstack{
			T-6
			\\
			\footnotesize{\textbf{Karel:Diagonal}~\cite{intro_to_karel_codehs}}
			}}
			&
			\multirow{2}{*}{
			\shortstack[c]{
			$8$
			\\
			\text{\scriptsize{\{\DSLRun\{\DSLWhile{}\}\}}}
			}
			}
			&
			\text{\scriptsize{\{\DSLRun{}}\}} 
			& $447$ & $4470$
			\\
			&&
			\ $S^{\text{in,}\star}$
			& $579$ & $5790$
			\\
		\bottomrule
   \end{tabular}
   }
	\caption{\looseness-1\algmultihop~applied to six HOC and Karel reference tasks; see Section~\ref{sec:evaluation} for details. For brevity, sketches have been abbreviated, e.g., \DSLRepeatUntil{}(\DSLBoolGoal) as \DSLRUntil.
    }
\label{fig:experiments.analysis}
\end{figure}

\section{\algmultihop~on Real-World Tasks}\label{sec:evaluation}
\looseness-1In this section, we present the performance of \algmultihop~on six reference tasks taken from real-world block-based programming platforms: HOC~\cite{hourofcode_maze} and Karel~\cite{intro_to_karel_codehs}. The set of these tasks along with their sources are mentioned in Fig.~\ref{fig:experiments.analysis}. These tasks differ in complexity, measured in terms of the programming constructs of their solution code as illustrated by the diversity of their respective solution sketches ${S^{\text{in,}\star}}$. For the exhaustive set of substructures of $S^{\text{in,}\star}$, 
Fig.~\ref{fig:experiments.analysis} lists the total number of pop quizzes, in the form of unique task-code pairs ($T^\text{quiz}, C^\text{quiz}$), generated by our algorithm.
As can be seen in the figure, our algorithm generates $50$ to $1000$s of pop quizzes for each substructure. For any potential student attempt on these tasks, Stage 1 of \algmultihop~ would generate one of these task-specific substructures by design -- hence, for every attempt we can present several unique yet adaptive pop quizzes to the student. Note that, our algorithm generates higher number of tasks than codes for each substructure. This is because the task synthesis methodology used in Stage 2(ii) can generate more than one task for a single code in Stage 2(ii) of Fig.~\ref{fig:pipeline.abstract}. In particular, for each new code, we obtain $10$ diverse tasks. For instance, Fig.~\ref{fig:intro} and Fig.~\ref{fig:karel.illustration} illustrate pop quizzes generated by \algmultihop{} for the specific student attempts on tasks T-4 and T-5, respectively.
%

%

\section{Expert Study via Multi-Dimensional Rubric
}\label{sec:surveystudy}

\looseness-1In this section, we evaluate \algmultihop~w.r.t. the desired features specified in the objective, i.e., \ind, \interp, and \conceal~(see Section~\ref{sec:notation.setup}). In particular, we seek to compare \algmultihop~with its variants resulting from different design choices in Section~\ref{sec:model}. To this end, we conduct an expert study via a multi-dimensional rubric.

\looseness-1\textbf{Variants of \algmultihop~algorithm.} We compare the performance of \algmultihop{} with the following variants: \algsame, \alghintpolicystruct, and \algmincode{}. \algsame~and \alghintpolicystruct~differ from \algmultihop~only in the \hintpolicy{} routine used in Stage 1  of Fig.~\ref{fig:pipeline.abstract} when generating $S^\text{quiz}$. In particular, Stage 1 of \algsame~always returns the sketch of the solution code, i.e., $S^\text{quiz} := S^{\text{in,}\star}$; Stage 1 of \alghintpolicystruct~returns a sketch directly from the $1$-hop neighborhood of $S^\text{in,stu}$, i.e., $S^\text{quiz} \in \mathcal{N}_\sketchspace(S^\text{in,stu},1)$. The third baseline, \algmincode, differs from \algmultihop~only in Stage 2(i) of Fig.~\ref{fig:pipeline.abstract} when generating $C^\text{quiz}$ from $S^\text{quiz}$. In particular, Stage 2(i) of \algmincode~generates $C^\text{quiz}$ as a direct reduction of the solution code w.r.t. the sketch obtained in Stage 1, i.e., $C^\text{quiz} \in \reducedcodes(\solutioncode~|~S^\text{quiz})$.

\looseness-1\textbf{Simulated student attempts.} For this expert evaluation, we simulated unsuccessful student attempts as seen in block-based programming domains~\cite{DBLP:conf/lats/PiechSHG15}. In particular, for each reference task, we manually created four student attempts as follows: (a) Stu-A: $C^\text{in,stu}$ uses only action blocks, i.e., (\DSLMove, \DSLTurnLeft, \DSLTurnRight, \DSLPickMarker, \DSLPutMarker); (b) Stu-B: $C^\text{in,stu}$ uses a subset of programming constructs in \solutioncode; (c) Stu-C: $C^\text{in,stu}$ is structurally the same as \solutioncode, i.e., $S^\text{in,stu}= S^{\text{in,}\star}$; (d) Stu-D: $C^\text{in,stu}$ has a structure more complex than \solutioncode. These four types of attempts exhaustively cover all the scenarios that an algorithm might encounter when deployed (see Section~\ref{sec:userstudy}).

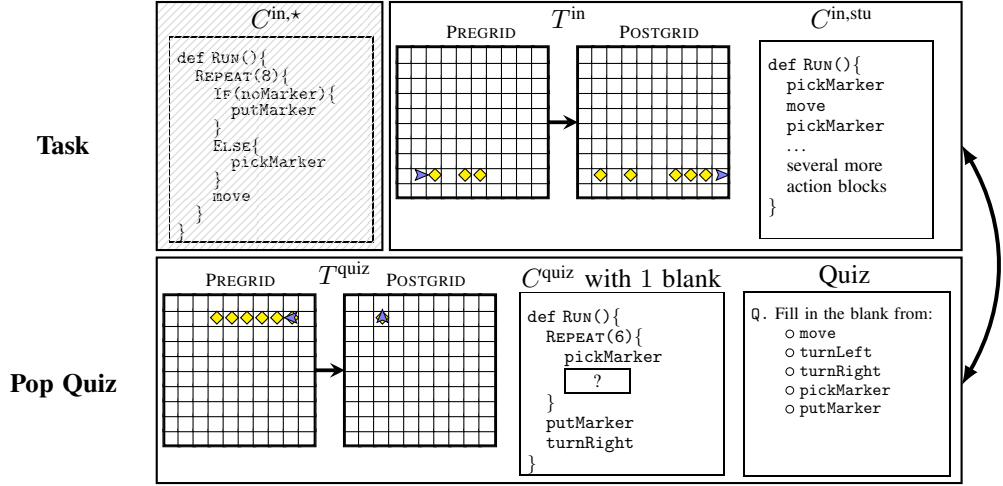
\begin{figure*}[!t]
\centering
\begin{tikzpicture}[
        box/.style={rectangle,draw=black, minimum size=0.1cm},scale=1, every node/.style={transform shape}
        ]
       
        \foreach \x in {-7.6,-7.4,...,-5.6}{
            \foreach \y in {1.9,2.1,...,3.9}
                \node[box, fill=white] at (\x,\y){};
        }

\foreach \x in {-5.2,-5,...,-3.4}{
            \foreach \y in {1.9,2.1,...,3.9}
                \node[box, fill=white] at (\x,\y){};
        }

\node[draw, fill=yellow, diamond, inner sep=1.25pt,minimum size=0.4pt] at(-7.2, 2.1){};
\node[draw, fill=yellow, diamond, inner sep=1.25pt,minimum size=0.4pt] at(-6.8, 2.1){};
\node[draw, fill=yellow, diamond, inner sep=1.25pt,minimum size=0.4pt] at(-6.6, 2.1){};
\node[draw, fill=yellow, diamond, inner sep=1.25pt,minimum size=0.4pt] at(-5, 2.1){};
\node[draw, fill=yellow, diamond, inner sep=1.25pt,minimum size=0.4pt] at(-4.6, 2.1){};
\node[draw, fill=yellow, diamond, inner sep=1.25pt,minimum size=0.4pt] at(-4, 2.1){};
\node[draw, fill=yellow, diamond, inner sep=1.25pt,minimum size=0.4pt] at(-3.8, 2.1){};
\node[draw, fill=yellow, diamond, inner sep=1.25pt,minimum size=0.4pt] at(-3.6, 2.1){};

\node[draw, fill=blue!50, dart, rotate=0 ,inner sep=0.01pt, minimum size=4pt] at (-7.4, 2.1){};
\node[draw, fill=blue!50, dart, rotate=0 ,inner sep=0.01pt, minimum size=4pt] at (-3.4, 2.1){};

\draw[draw=black, thick] (-7.7,1.8) rectangle (-5.7,3.8);
\draw[draw=black, thick] (-5.3,1.8) rectangle (-3.3,3.8);
\draw[-stealth, line width=0.5mm] (-5.7, 2.8) -- (-5.3,2.8);
        \node[text width=0.5cm, anchor=west, right] at (-3.25, 2.85)
    {\begin{boxcode}{2.98cm}{0.70}{1.0}
				\textcode{def }\DSLRun\textcode{()\{}\\
                \quad \DSLPickMarker\\
                \quad \DSLMove\\
                \quad \DSLPickMarker\\
                \quad \ldots\\
                \quad \text{several more} \\ \quad \text{action blocks}\\
				\textcode{\}}
				\\
			\end{boxcode}};
	\node[text width=0.5cm, anchor=east, left] at (-10.35, 2.85)
    {\begin{boxcode}{3.68cm}{0.68}{0.8}
				\textcode{def }\DSLRun\textcode{()\{}\\
				\quad \DSLRepeat\textcode{(8}\textcode{)\{}\\
                \quad \quad
              \DSLIf\textcode{(}\DSLBoolNoMarker\textcode{)\{}\\
				\quad \quad \quad \DSLPutMarker\\
				\quad \quad \textcode{\}}\\
                \quad \quad \DSLElse\textcode{}\textcode{\{}\\
                \quad \quad \quad \DSLPickMarker\\
                \quad \quad \textcode{\}
                	}\\
                \quad \quad \DSLMove\\
				\quad \textcode{\}}\\
				\textcode{\}}
			\end{boxcode}
			};
	\draw[draw=black, thick, pattern=north east lines, pattern color=gray!30] (-10.9,1.1) rectangle (-7.9, 4.4);
	\node[text width=0.5cm] at (-5.4,4.2) {$T^\text{in}$};
	\node[text width=0.5cm] at (-6.8,4) {\scriptsize\textsc{{Pregrid}}};
	\node[text width=0.5cm] at (-4.5,4) {\scriptsize\textsc{{Postgrid}}};
	\node[text width=0.5cm] at (-1.96,4.2) {\studentcode};
	\node[text width=0.5cm] at (-9.4,4.2) {\solutioncode};
    \draw[draw=black, thick] (-7.8,1.1) rectangle (-0.2, 4.4);
	\draw[draw=black, thick] (-10.9,-2) rectangle (-0.2, 1);
[
        box/.style={rectangle,draw=black, minimum size=0.2cm},
        ]
        
        \foreach \x in {-10.7,-10.5,...,-8.9}{
            \foreach \y in {-1.4,-1.2,...,0.4}
                \node[box, fill=white] at (\x,\y){};
        }

\foreach \x in {-8.3,-8.1,...,-6.5}{
             \foreach \y in {-1.4,-1.2,...,0.4}
                \node[box, fill=white] at (\x,\y){};
        }

\node[draw, fill=yellow, diamond, inner sep=1.25pt,minimum size=0.5pt] at(-10.1, 0.2){};
\node[draw, fill=yellow, diamond, inner sep=1.25pt,minimum size=0.5pt] at(-9.9, 0.2){};
\node[draw, fill=yellow, diamond, inner sep=1.25pt,minimum size=0.5pt] at(-9.7, 0.2){};
\node[draw, fill=yellow, diamond, inner sep=1.25pt,minimum size=0.5pt] at(-9.5, 0.2){};
\node[draw, fill=yellow, diamond, inner sep=1.25pt,minimum size=0.5pt] at(-9.3, 0.2){};
\node[draw, fill=yellow, diamond, inner sep=1.25pt,minimum size=0.5pt] at(-9.1, 0.2){};
\node[draw, fill=yellow, diamond, inner sep=1.25pt,minimum size=0.5pt] at(-7.9, 0.2){};

\node[draw, fill=blue!50, dart, rotate=-180 ,inner sep=0.01pt, minimum size=4pt] at (-9.1, 0.2){};
\node[draw, fill=blue!50, dart, rotate=90 ,inner sep=0.01pt, minimum size=4pt] at (-7.9, 0.2){};

\draw[draw=black, thick] (-10.8,-1.5) rectangle (-8.8,0.5);
\draw[draw=black, thick] (-8.4,-1.5) rectangle (-6.4,0.5);
\draw[-stealth, line width=0.5mm] (-8.8, -0.5) -- (-8.4,-0.5);
 
    \node[text width=0.5cm, anchor=west, right] at (-6.45, -0.4)
    {\begin{boxcode}{3.58cm}{0.70}{1.0}
				\textcode{def }\DSLRun\textcode{()\{}\\
				\quad \DSLRepeat\textcode{(6}\textcode{)\{}\\
                \quad \quad \DSLPickMarker\\
                \quad \quad \framebox[8.0\width]{?}\\
				\quad \textcode{\}}\\
				\quad \DSLPutMarker\\
				\quad \DSLTurnRight\\
				\textcode{\}}
			\end{boxcode}};
	 \node[text width=0.5cm, anchor=west, right] at (-3.45, -0.4)
	 {\begin{boxcode}{3.92cm}{0.65}{1}
				\textcode{Q.} \text{Fill in the blank from: }\\
				\quad \quad \tikz\draw[black,fill=none] (0,0) circle (.5ex); \DSLMove\\
				\quad \quad \tikz\draw[black,fill=none] (0,0) circle (.5ex); \DSLTurnLeft\\
				\quad \quad \tikz\draw[black,fill=none] (0,0) circle (.5ex); \DSLTurnRight\\
				\quad \quad \tikz\draw[black,fill=none] (0,0) circle (.5ex); \DSLPickMarker\\
				\quad \quad \tikz\draw[black,fill=none] (0,0) circle (.5ex); \DSLPutMarker
			\\
			\\
			\\
			\end{boxcode}
	 };
\node[text width=0.5cm] at (-8.5,0.8) {$T^\text{quiz}$};
\node[text width=0.5cm] at (-10,0.7) {\scriptsize\textsc{{Pregrid}}};
	\node[text width=0.5cm] at (-7.6,0.7) {\scriptsize\textsc{{Postgrid}}};
\node[text width=2.6cm] at (-4.75,0.75) {$C^\text{quiz}\text{ with }1\text{ blank}$};
\node[text width=1cm] at (-1.6,0.75) {Quiz};

\tikzstyle{doublearr}=[latex-latex, black, line width=1.5pt]
\draw [doublearr, bend left]    (-0.2, 2.5) to (-0.2,-0.7);

    \node[text width=1cm] at (-12,2.5){\textbf{Task}};
    \node[text width=1.7cm] at (-12,-0.7){\textbf{Pop Quiz}};
    \end{tikzpicture}
\caption{Analogous to Fig.~\ref{fig:intro}, here we illustrate our framework on a Karel task, T-5 (see Fig.~\ref{fig:experiments.analysis}). Karel tasks~\cite{pattis1981karel} comprise of a pair of visual grids, (\textsc{Pregrid}, \textsc{Postgrid}), and the objective is to write code that, when executed, transforms \textsc{Pregrid} to \textsc{Postgrid}. 
}
\label{fig:karel.illustration}
\end{figure*}

\looseness-1\textbf{Multi-dimensional evaluation rubric.} Inspired by the evaluation rubric in \cite{DBLP:conf/aied/PriceZB17,DBLP:conf/edm/ZhiMDLPB19},  we assess pop quizzes on a multi-dimensional rubric with three attributes, each rated on a three-point Likert scale (with higher scores being better). More concretely, we have: (i) \ind~attribute measuring the degree of individualization of the pop quiz to the current student attempt ($3$: high; $2$: medium; $1$: low); (ii) \interp~attribute measuring how easy the pop quiz is to comprehend/solve ($3$: easy; $2$: might confuse the student sometimes; $1$: either incorrect or is very difficult to solve.); (iii) \conceal~attribute measuring the extent to which the pop quiz conceals the solution code ($3$: sufficiently conceals; $2$: reveals the solution to some extent; $1$: reveals the solution to a large extent). \overall{} denotes the sum of scores across three attributes for a pop quiz.

\looseness-1\textbf{Expert study setup.} We picked three tasks spanning different types of constructs and complexity: T-1, T-4, and T-5 from Fig.~\ref{fig:experiments.analysis}. Thus, in total we evaluated $48$ scenarios: $4$ algorithm variants $\times$ $4$ student types $\times$ $3$ tasks (see Figs.~\ref{fig:intro} and~\ref{fig:karel.illustration} as example scenarios). Two researchers, with experience in block-based programming, evaluated each of the $48$ scenarios independently.
The evaluation was done through a web survey where a scenario was introduced at random, and assessed based on the rubric.


\begin{wrapfigure}[11]{r}{0.62\textwidth}
\vspace{-3.5mm}
\centering
\scalebox{0.85}{
    \setlength\tabcolsep{2pt}
    \renewcommand{\arraystretch}{1.1}
    \begin{tabular}{c|ccc|c}
        Algorithm & \small{\ind} & \small{\interp} & \small{\conceal} & \small{\overall} \\
        \toprule
        \footnotesize{\algsame} & \cellcolor{red!35}$2.0 (0.7)$ & $2.8 (0.1)$ & $3.0 (0.0)$ & $7.8 (0.8)$ \\
        \footnotesize{\alghintpolicystruct} & $2.8 (0.1)$ & \cellcolor{red!35}$2.5 (0.6)$ & $3.0 (0.0)$ & $8.3 (0.7)$ \\
        \footnotesize{\algmincode} & $2.7 (0.3)$ & $3.0 (0.0)$ & \cellcolor{red!35}$1.5 (0.4)$ & $7.2 (0.7)$ \\
        \hline
        \cellcolor{blue!15}\footnotesize{\algmultihop} & \cellcolor{blue!15}${2.7 (0.2)}$ & \cellcolor{blue!15}${3.0 (0.0)}$ & \cellcolor{blue!15}${2.9 (0.1)}$ & 
        \cellcolor{blue!15}${8.6 (0.3)}$ \\
        \bottomrule
    \end{tabular}
}
\caption{\looseness-1Mean (Variance) attribute ratings for different algorithms. Higher scores are better. \algmultihop~performs well across all three attributes and has the highest \overall~score; 
see  Section~\ref{sec:surveystudy} for details.}
\label{fig:surveystudy}
\end{wrapfigure}
%


\looseness-1\textbf{Expert study results.} First, we validate the expert ratings using the quadratic-weighted Cohen's kappa inter-agreement reliability value~\cite{DBLP:conf/aied/PriceZB17} for each attribute: $0.62$ (\ind), $0.69$ (\interp), $0.79$ (\conceal), and $0.7$ (\overall). The values indicate \textit{substantial agreement} between the raters. The average ratings are presented in Fig.~\ref{fig:surveystudy} and \algmultihop~has the highest \overall~score. We analyze these ratings per attribute based on the Kruskal-Wallis significance test~\cite{macfarland2016kruskal}; the results discussed next are statistically significant with $p < 0.01$. On the~\ind~attribute, \algsame{} performs significantly worse because it does not account for the student attempt (see Section~\ref{sec:model.stage1}). On the ~\interp~attribute, \alghintpolicystruct~performs significantly worse because there are instances where no valid code reduction of \solutioncode~w.r.t. $S^\text{quiz}$ is found (see Footnote~\ref{footnote:sec3}, Section~\ref{sec:model.stage2}). 
Finally, on the \conceal~attribute, \algmincode~performs significantly worse because it obtains $C^\text{quiz}$ via a direct reduction of \solutioncode{} without any mutation (see Section~\ref{sec:model.stage2}). 

\section{User Study via Online Platform}
\label{sec:userstudy}
\looseness-1We have built an online platform with our \algmultihop~framework using the Blockly Games library~\cite{blocklygames}. The online platform is publicly accessible -- see Footnote~\ref{footnote:webapp}, Section~\ref{sec:intro.contributions}. The platform provides an interface for a participant to practice block-based programming tasks, and receive pop quiz based feedback when stuck. In this section, we report results from an initial user study to assess the efficacy of our scaffolding framework in comparison to other feedback methods. 
%

\looseness-1\textbf{Participation session and feedback methods.} A single  session on our platform comprises of three steps. In \textsc{Step-A}, the participant is presented with a task and has $10$  execution tries to solve it. If a participant fails to solve the task at \textsc{Step-A}, they proceed to \textsc{Step-B} with a randomly assigned feedback method (\algnohint, \algactionhint, and \algmultihop~as discussed below). After \textsc{Step-B}, the participant resumes their attempt on the task in \textsc{Step-C} with $10$ additional execution tries. Note that the feedback method is invoked only once in a single session. Next, we describe different feedback methods at \textsc{Step-B}. \algnohint{} represents a baseline where the participant is directed to \textsc{Step-C} without any  feedback.  \algactionhint{} corresponds to next-step hints as feedback where the participant's code is updated to bring it closer to a solution code~\cite{DBLP:conf/its/RiversK14,DBLP:conf/lats/PiechSHG15,DBLP:journals/aiedu/RiversK17,DBLP:conf/kbse/ZimmermanR15,DBLP:conf/edm/PriceZB17}; we prioritized next-step edits involving programming constructs (e.g., \DSLRepeatUntil) over basic actions (e.g., \DSLMove). \algmultihop~is our pop quiz based feedback.


\begin{wrapfigure}[14]{r}{0.61\textwidth}
\vspace{-3mm}
\centering
\scalebox{0.85}{
    \setlength\tabcolsep{3pt}
    \renewcommand{\arraystretch}{1.2}
    \begin{tabular}{r|rrr|rrr}
			\toprule
			Feedback & \multicolumn{3}{c|}{Total (\textbf{\textsc{Step-B}})} & \multicolumn{3}{c}{Fraction solved (\textbf{\textsc{Step-C})}} \\
			 & Both & T-3 & T-5 & Both & \ \ \ T-3 & T-5\\
			\toprule
			\algnohint & $151$ & $63$ & $88$ & $0.046$ & \ \ \  $0.079$ & $0.023$ \\
			\algactionhint & $146$ & $63$ & $83$ & $0.082$ & \ \ \ $0.127$ & $0.048$ \\
			\hline
			\cellcolor{blue!15}\algmultihop & \cellcolor{blue!15}$148$ & \cellcolor{blue!15}$62$ & \cellcolor{blue!15}$86$ & \cellcolor{blue!15}$0.128$ & \ \ \ \cellcolor{blue!15}$0.177$ & 
			\cellcolor{blue!15}$0.093$ \\
			\bottomrule
   \end{tabular}
}
\caption{\looseness-1Results for tasks T-3 and T-5 (``Both'' represents aggregated results). In \textbf{\textsc{Step-A}}, we had a total of $575$ ($293$ for T-3, $282$ for T-5) participants; about $0.774$ ($0.642$ for T-3, $0.911$ for T-5) fraction failed to solve the task at \textbf{\textsc{Step-A}} and proceeded to \textbf{\textsc{Step-B}} / \textbf{\textsc{Step-C}} with a randomly assigned feedback method.}
\label{table:userstudy}
\end{wrapfigure}

\looseness-1\textbf{User study results.} We conducted an initial user study with participants recruited from Amazon Mechanical Turk; an IRB approval was received before the study. The participants were US-based adults, without expertise in block-based visual programming. Due to the costs involved (over $3$ USD per participant), we selected two tasks for the study:  T-3 and T-5 from Fig.~\ref{fig:experiments.analysis}. We present the detailed results in Fig.~\ref{table:userstudy}. In total, we had $575$ unique participants; out of these, $0.774$ fraction failed to solve the task at \textsc{Step-A} and proceeded to \textsc{Step-B}. \algmultihop{} was assigned to $148$ participants in \textsc{Step-B} ($0.60$ fraction successfully solved the presented pop quiz). Subsequently, $0.128$ fraction of these participants solved the task in \textsc{Step-C}. Here, $0.128$ measures the success rate of participants assigned to \algmultihop{}; in comparison, it is $0.082$ for \algactionhint{}  and $0.046$ for \algnohint{} -- see Fig.~\ref{table:userstudy}.
%
Overall, the performance of \algmultihop{} is better than \algactionhint{} (the gap is not significant w.r.t. $\chi^2$-test, $p = 0.19$) and \algnohint{} (the gap is significant w.r.t. $\chi^2$-test, $p = 0.01$)~\cite{mchugh2013chi}. These initial results demonstrate the utility of providing pop quiz based feedback.



\section{Conclusions and Outlook}
\label{sec:conclusions}
\looseness-1We proposed a novel scaffolding framework for block-based programming based on pop quizzes that involve inquiry-driven and problem-solving aspects. We developed a modular synthesis algorithm, \algmultihop, that generates these pop quizzes. After conducting an expert assessment using a multi-dimensional rubric, we developed an online platform empowered by our scaffolding framework. While initial user study results with our platform demonstrate the utility of our pop quiz based framework, there are several interesting directions to continue this study, including: (i) extending our platform to provide multiple rounds of feedback within a single participation session and measuring the efficacy of different methods; (ii) comparing our synthesized pop quizzes with those generated by experts; (iii) conducting longitudinal studies with novice students to measure long-term improvements in problem solving skills; (iv) extending our framework to more complex block-based programming domains.


\looseness-1\textbf{Acknowledgments.} We would like to thank the reviewers for their feedback. Ahana Ghosh acknowledges support by Microsoft Research through its PhD Scholarship Programme.
%
%
Funded/Co-funded by the European Union (ERC, TOPS, 101039090). Views and opinions expressed are however those of the author(s) only and do not necessarily reflect those of the European Union or the European Research Council. Neither the European Union nor the granting authority can be held responsible for them.

\clearpage
\bibliographystyle{plain}
\bibliography{main}
\end{document}